\documentclass[runningheads]{llncs}
\usepackage{graphicx}

\usepackage{tikz}
\usepackage{comment}
\usepackage{amsmath,amssymb} 
\usepackage{color}
\usepackage{hyperref}
\usepackage{orcidlink}

\usepackage{caption}
\usepackage{subfigure}
\usepackage{floatrow}
\newfloatcommand{capbtabbox}{table}[][\FBwidth]

\newcommand{\etal}{\textit{et al}.}
\newcommand{\ie}{\textit{i}.\textit{e}.}
\newcommand{\eg}{\textit{e}.\textit{g}.}
\newcommand*\samethanks[1][\value{footnote}]{\footnotemark[#1]}

\begin{document}
\pagestyle{headings}
\mainmatter
\def\ECCVSubNumber{2328}  

\title{Mining Cross-Person Cues for Body-Part Interactiveness Learning in HOI Detection} 

\titlerunning{ECCV-22 submission ID \ECCVSubNumber} 
\authorrunning{ECCV-22 submission ID \ECCVSubNumber} 
\author{Anonymous ECCV submission}
\institute{Paper ID \ECCVSubNumber}

\titlerunning{Mining Cross-Person Cues for Body-Part Interactiveness Learning}
%
\author{Xiaoqian Wu\inst{1}\orcidlink{0000-0003-1566-3811}\thanks{The first two authors contribute equally.} \and
Yong-Lu Li\inst{1,2}\orcidlink{0000-0003-0478-0692}\samethanks \and
Xinpeng Liu \inst{1}\orcidlink{0000-0002-7525-3243} \and
Junyi Zhang \inst{1}\orcidlink{0000-0002-9291-3098} \and
Yuzhe Wu \inst{3}\orcidlink{0000-0003-4603-5295} \and
Cewu Lu\inst{1}\orcidlink{0000-0002-4023-9257}\thanks{Cewu Lu is corresponding author,  member of Qing Yuan Research Institute and Shanghai Qi Zhi institute.}
}
\authorrunning{X. Wu et al.}
%
\institute{Shanghai Jiao Tong University 
\and
Hong Kong University of Science and Technology
\and
DongHua University\\
\email{\{enlighten,yonglu\_li,junyizhang,lucewu\}@sjtu.edu.cn, \{xinpengliu0907,wuyuzhe486\}@gmail.com}
}
\maketitle

\begin{abstract}
Human-Object Interaction (HOI) detection plays a crucial role in activity understanding. Though significant progress has been made, interactiveness learning remains a challenging problem in HOI detection: existing methods usually generate redundant negative H-O pair proposals and fail to effectively extract interactive pairs. 
Though interactiveness has been studied in both whole body- and part- level and facilitates the H-O pairing, previous works only focus on the \textit{target person} once 
(\textit{i.e.}, in a \textbf{local} perspective) and overlook the information of the other persons.
In this paper, we argue that comparing body-parts of multi-person simultaneously can afford us more useful and supplementary interactiveness cues.
That said, to learn body-part interactiveness from a \textbf{global} perspective: when classifying a target person’s body-part interactiveness, visual cues are explored not only from herself/himself but also from \textit{other persons in the image}.
We construct body-part saliency maps based on self-attention to mine cross-person informative cues and learn the holistic relationships between \textit{all} the body-parts. 
We evaluate the proposed method on widely-used benchmarks HICO-DET and V-COCO. With our new perspective, the holistic global-local body-part interactiveness learning
achieves significant improvements over state-of-the-art. 
Our code is available at \href{https://github.com/enlighten0707/Body-Part-Map-for-Interactiveness}{https://github.com/enlighten0707/Body-Part-Map-for-Interactiveness}.
\end{abstract}

\section{Introduction}\label{sec:intro}

Human-Object Interaction (HOI) detection retrieves human and object locations and infers the interactions simultaneously from still images. In practice, an HOI instance is represented as a triplet: (\textit{human, verb, object}). As a sub-task of visual relationship~\cite{visualgenome,Lu2016Visual}, 
it is crucial for activity understanding, embodied AI, etc. 

\begin{figure}
\centering
\includegraphics[width=\textwidth]{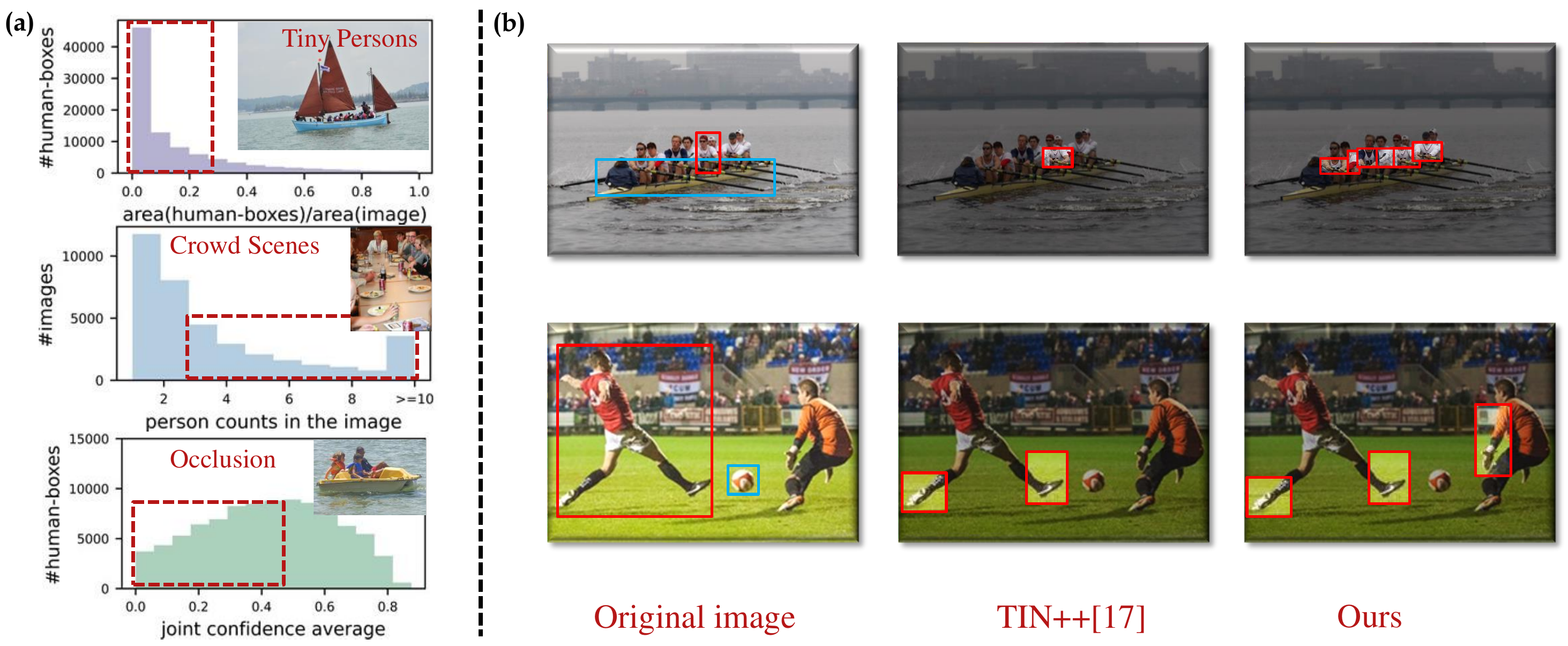}
\caption{
a) Statistics of hard cases in HOI datasets HICO-DET~\cite{hicodet}. 
The images containing tiny persons, crowded scenes, and occlusion are considered.
b) Our idea of learning body-part interactiveness from the same/different body-parts of other persons in the image.
}
\label{fig:intro} 
\end{figure}

Though significant progress has been made, HOI detection is still bottlenecked by interactiveness learning~\cite{li2019transferable}: they fail to effectively extract interacted pairs while generating redundant negative pairs. 
The problem is first raised in TIN~\cite{li2019transferable}, where a pairwise interactiveness classifier is inserted into the HOI detection framework and the interactiveness prediction is used for non-interaction suppression. The decent gain on HOI detection performance verifies the great potential and importance of interactiveness learning.

Recently, TIN++~\cite{li2021transferable} is further proposed to utilize body-part level features to improve instance-level interactiveness learning via a hierarchical diagram. Despite the improvement in detecting positive pairs, it focuses on \textit{local} body-part features from the targeted person only, which is not enough. We argue that when classifying the target person's body-part interactiveness, visual cues can be explored not only from himself/herself but also from \textit{other persons in the image}. 
First, aside from the targeted person and object, it is also important to exploit contextual cues from the whole image~\cite{gao2018ican}. 
Existing methods~\cite{gao2018ican,qpic} have made efforts to learn relevant context to facilitate HOI learning.
However, it is hard and \textbf{unstable} to retrieve useful information from various backgrounds without restriction. Therefore, we argue to better utilize contextual information by highlighting \textit{all the persons} in the image because regions containing persons are usually more informative according to our prior. For instance, when recognizing a speaker giving a lecture, the audience is evidence more obviously than the stage decoration.
Furthermore, as is shown in Fig.~\ref{fig:intro}a, there are a large proportion of \textbf{hard cases} in HOI datasets~\cite{hicodet}, \eg, tiny interactive persons, crowded scenes, and occlusion. In these cases, the cue carried by body parts of the local targeted person is very limited,
while mining cues from a \textit{global} multi-person perspective with other persons as a reference would be a good choice to alleviate the difficulty. 

Following this insight, we propose to learn body-part interactiveness from the same/different body-parts of other persons in the image, which is illustrated in Fig.~\ref{fig:intro}b.
In the upper image, a crowd of tiny persons is rowing boats, and their lower bodies are occluded. When classifying a targeted H-O pair, 
aside from the targeted object, 
TIN++~\cite{li2021transferable} only focuses on the targeted person from the body-part level (\eg, hand), which is highlighted in the image. 
Nevertheless, we also emphasize other persons' hands even when classifying the targeted person's hand interactiveness, which provides a \textbf{supplementary} and \textbf{contrastive} viewpoint for interactiveness learning. 
Suppose another person B's hands are given as a reference which is easier to identify interactiveness, then the similarity between B's and targeted person A's hands will lead to the same prediction (both interactive or non-interactive), while discernable difference will lead to an opposite conclusion.
Further, attention to different body-parts of multi-person also matters. In the bottom image, the left person is kicking a football while the right person is defending. When classifying the left person's \textit{feet} interactiveness, the right person's \textit{arms} would provide useful cues, since he is stretching out his arms and trying to intercept the ball. Thus, the relationship between \textit{different} body-parts of different persons overlooked by previous works also offers supplementary information to interactiveness inference.

In light of these, we utilize a transformer for body-part interactiveness detection, where self-attention helps to capture informative cross-person visual cues.
\textbf{First}, \textit{body-part saliency maps} are constructed via image patches (\ie, transformer tokens) masking, where patches not containing interested body-parts are masked and blocked from the  computation. 
\textbf{Second}, to encode diverse visual patterns more flexibly, body-parts are \textit{progressively masked}, where different attention mask is applied in successive transformer layers and more tokens are dropped in the late layers. 
\textbf{Third}, motivated by the sparsity property~\cite{hake,li2021transferable} of body-part interactiveness~\cite{li2021transferable}, the model classifies interactiveness of different body-parts via \textit{one-time passing} to improve computation efficiency. An early filter is inserted to drop unimportant body-parts, and then the remaining saliency maps are merged.
\textbf{Fourth}, 
we also propose a \textit{sparsity adaptive sampling strategy} on the train set to put more emphasis on crowded scenes and guide better interactiveness knowledge.
In extensive experiments, the proposed method achieves state-of-the-art.
We firstly achieve all \textbf{33+} mAP on three sets of HICO-DET~\cite{hicodet}, especially the impressive improvement on Rare HOIs (\textbf{6.16} mAP improvement upon the SOTA CDN~\cite{cdn}) thanks to our holistic global-local interactiveness learning. 
Meanwhile, on the HOI hard cases, we also show our significant superiority.

Our contribution includes: 
1) We propose to learn body-part interactiveness from a global perspective as an effective supplement for existing local-based methods, thus boosting interactiveness learning; 
2) To mine cross-person cues,
we construct body-part saliency maps based on self-attention computation and propose the progressively mask and one-time passing strategies to 
improve flexibility and efficiency;
3) With our proposed interactiveness detector, we achieve state-of-the-art on widely-used benchmarks HICO-DET~\cite{hicodet} and V-COCO~\cite{vcoco}.

\section{Related Work}
\noindent{\bf Human-Object Interaction.}
Human-Object Interaction is essential to understand human-centric interaction with objects. Rapid progress has been made in HOI learning. Many large datasets~\cite{hicodet,vcoco,hake,openimages} and deep learning based methods~\cite{hicodet,qi2018learning,Gkioxari2017Detecting,gao2018ican,li2019transferable,li2021transferable,djrn,vcl,li2020hoi,ppdm,ipnet,qpic,cdn,liu2022highlighting,liu2022interactiveness} have been proposed. They usually followed the two-stage pipeline, \textit{i.e.}, first H-O pair detection, and then HOI classification.
Chao \etal~\cite{hicodet} proposed the widely-used multi-stream framework combining visual features and spatial locations, while GPNN~\cite{qi2018learning} incorporated DNN and graph model to model the HOI relationship. InteractNet~\cite{Gkioxari2017Detecting} utilized an action-specific density map to estimate the interacted object locations. 
DJ-RN~\cite{djrn} introduced 3D information and proposed a 2D-3D joint representation learning method. 
PaStaNet~\cite{hake} inferred human part states~\cite{lu2018beyond} first and then reason out the activities based on  part-level semantics. 
IDN~\cite{li2020hoi} analyzed HOI from an integration and decomposition perspective.
Recently, several one-stage methods have been proposed~\cite{ppdm,ipnet,qpic,cdn}, where HOIs triplets are directly detected by parallel HOI detectors.
PPDM~\cite{ppdm} and IP-Net~\cite{ipnet} adopted a variant of one-stage object detector for HOI detection. 
QPIC~\cite{qpic} utilized the recent transformer-based object detector DETR~\cite{detr} to aggregate image-wide contextual information and facilitate HOI learning. 

\noindent{\bf Interactiveness Learning.}
Though significant progress has been made, interactiveness learning remains challenging in HOI detection. Existing methods fail to pair interactive human and object effectively but generate redundant negative pairs. 
TIN and TIN++~\cite{li2019transferable,li2021transferable} 
first raised this problem and tried to address it via an inserted pairwise interactiveness classifier. In TIN++~\cite{li2021transferable}, the framework is extended to a hierarchical paradigm with jointly learning instance-level and body part-level interactiveness.
Recently, CDN~\cite{cdn} proposed to accurately locate interactive pairs via a one-stage transformer framework disentangling human-object pair detection and interaction classification. However, it still performs not well on interactiveness detection (Sec.~\ref{sec:result}).
Here, we point out the previously overlooked global perspective and utilize it to improve interactiveness learning.

\noindent{\bf Part-based Action Recognition.}
The part-level human feature provides finer-grained visual cues to improve HOI detection.  
Based on the whole person and part boxes, Gkioxari \etal~\cite{gkioxari2015actions} developed a part-based model to make fine-grained action recognition. Fang \etal~\cite{Fang2018Pairwise} proposed a pairwise body-part attention model which can learn to focus on crucial parts and their correlations, while TIN++~\cite{li2021transferable} utilized the human instance and body-part features together to learn interactiveness in a hierarchical paradigm and extract deeper interactive visual clues.

\section{Method}
\subsection{Overview}
\begin{figure}
\centering
\includegraphics[width=\textwidth]{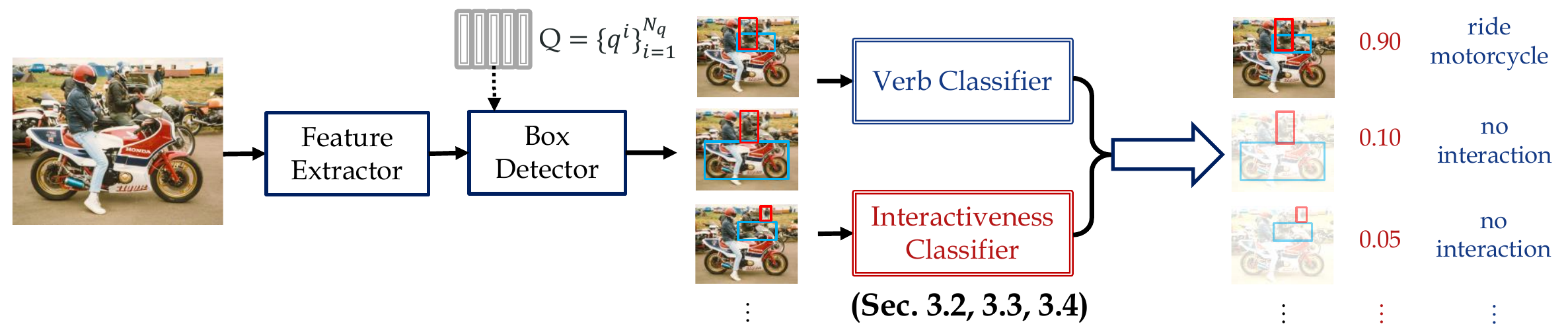}
\caption{The overall framework of our proposed method. 
} 
\label{fig:overview} 
\end{figure}

As is shown in Fig.~\ref{fig:overview}, our pipeline consists of three main modules: a \textit{box detector}, an \textit{interactiveness classifier}, and a \textit{verb classifier}. They are all implemented as stacked transformed decoder.

Given an input image $x\in \mathcal{R}^{3\times H_0 \times W_0}$, we first adopt a ResNet-50 followed by a transformer encoder as our visual feature extractor. The output feature map is obtained as $z \in \mathcal{R}^{D_c \times H \times W}$, where $D_c$ is the number of channels and $H, W$ is the size. A fixed positional embedding $pos \in \mathcal{R}^{D_c \times H \times W}$ is additionally input to the transformer to supplement the positional information. 
Then based on the feature map $z$, the three main components are used for HOI detection. 

An interactive human-object pair is mathematically defined following~\cite{qpic}: 
\textbf{1)} a human bounding box vector $b(h) \in [0,1]^4$ normalized by the image size, 
\textbf{2)} an object bounding box vector $b(o) \in [0,1]^4$ normalized by the image size, 
\textbf{3)} an object class vector $c \in \{0,1\}^{N_{obj}}$, where $N_{obj}$ is the number of object classes, 
\textbf{4)} an interactiveness prediction $p_{int} \in [0,1]$, and
\textbf{5)} a verb prediction $p_{verb} \in \{0,1\}^{N_v}$, where $N_v$ is the number of verb classes. 
For \textbf{box detection}, a transformer decoder $f_{dec1}(\cdot, \cdot, \cdot)$ transforms a set of learnable query vectors $Q=\{q^i|q^i \in \mathcal{R}^{D_c}\}_{i=1}^{N_q}$ into a set of decoded embeddings $D=\{d^{i}| d^{i} \in \mathcal{R}^{D_c}\}_{i=1}^{N_q}$, which is obtained as $D = f_{dec1}(z, Q, pos)$.
The subsequent three small feed-forward networks (FFNs): human bounding box FFN $f_h$, object bounding box FFN $f_o$, and object class FFN $f_c$ further process the decoded embeddings $D$ to produce $N_q$ prediction results $\{b(h)^i\}_{i=1}^{N_q}$, $\{b(o)^i\}_{i=1}^{N_q}$, $\{c^i\}_{i=1}^{N_q}$, respectively. The decoded embeddings $D$ are then fed into the interactiveness classifier and verb classifier as their query embeddings.
For \textbf{verb classification}, another transformer decoder $f_{dec3}(\cdot, \cdot, \cdot)$ 
takes $D$ as input and outputs $V=\{v^i| v^i \in \mathcal{R}^{D_c}\}_{i=1}^{N_q}$. With the verb class FFN $f_v$, the classification results is obtained as $p_{verb}^i = Sigmoid(f_v(v^i))$.

\begin{figure}
\centering
\includegraphics[width=\textwidth]{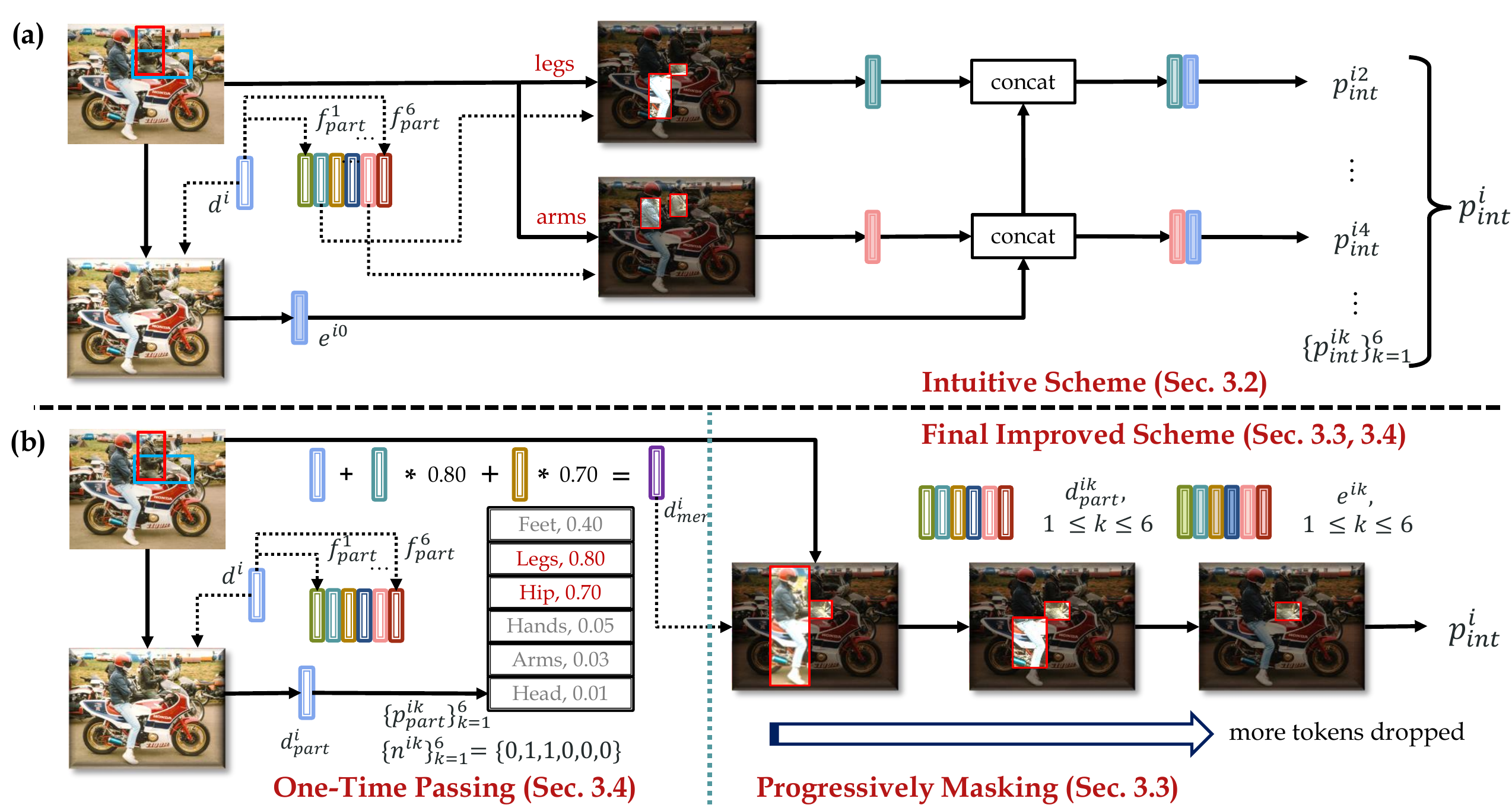}
\caption{The design of interactiveness classifier. 
a) Body-part saliency map construction and the intuitive scheme. We use legs and arms as examples.
b) The final improved model.
Here, we use the images to show the masking process for clarity instead of masking real images. The decoded embedding $d^i$ is generated from the box detector and fed into the interactiveness classifier as query embeddings.
}
\label{fig:module}
\end{figure}

Our main contribution is global-local interactiveness learning based on the proposed body-part saliency map. Thus, we focus on the design of the new proposed \textbf{interactiveness classifier} (Fig.~\ref{fig:module}).
In Sec.~\ref{sec:intuitive}, we introduce the construction of body-part saliency map based on self-attention computation, and provide an intuitive scheme to validate its effectiveness (Fig.~\ref{fig:module}a).
Then, we improve the intuitive scheme from two aspects: progressively masking to encode diverse visual patterns flexibly (Sec.~\ref{sec:progressive}), and one-time passing to improve efficiency (Sec.~\ref{sec:filter&merge}). The final improved model is shown in Fig.~\ref{fig:module}b.

\subsection{Constructing Body-Part Saliency Map}
\label{sec:intuitive}
We divide a person into six body-parts: feet, legs, hip, hands, arms, and head following HAKE~\cite{li2019hake}. To construct the body-part saliency map, \eg, for hands, we utilize the attention mask matrix in the transformer. Only image patches (or equally, transformer tokens) containing hands are remained for attention calculation, while other patches are masked. 
The attention mask is pre-calculated from body-part boxes detection results.

\subsubsection{An Intuitive Scheme}
In implementation, a transformer decoder $f_{dec2}(\cdot, \cdot, \cdot, \cdot)$ is used to transform input query embeddings $D$ into $E=\{e^i| e^i \in \mathcal{R}^{D_c}\}_{i=1}^{N_q}$. Body-part saliency map is applied in $f_{dec2}$ by masking partial of the feature map $z$ via a masking matrix $M = \{m^i | m^i \in \{0,1\}^{H \times W}\}_{i=1}^{N_q}$ (1 for effective tokens and 0 for masked tokens). 
First, to integrate image-wide contextual information, a set of decoded embeddings are obtained as $E^0 = f_{dec2}(z, D, pos, M^0)$, where $M^0$ is an all-one matrix, \ie, applying no mask. 
Next, the finer-grained body-part level features are used. The input query embeddings $D_{part}^{k}$ for the $k$-th ($1 \le k \le 6$) body-part is transformed from the original query embeddings $D$ via FFNs $\{f_{part}^{k}\}_{k=1}^6$, \ie, $D_{part}^{k}= f_{part}^{k}(D)$. Then the decoded embeddings are calculated via $E^k = f_{dec2}(z, D^{k}_{part}, pos, M^{k})$, where $M^{k}$ is the corresponding mask matrix. In $f_{dec2}$, the attention masks are different for each body-part, while the learned parameters are shared.
Finally, based on decoded embeddings $E^0$ and $\{E^k\}_{i=1}^{6}$, an interactiveness FFN $f_{int}$ is used to generate body-part interactiveness prediction. For the $i$-th proposal and the $k$-th body-part, its interactiveness score is obtained via $p_{int}^{ik} = Sigmoid(f_{int}(concat(e^{ik}, e^{i0})))$. The instance-level interactiveness score is then obtained as $p_{int}^{i} = max_k\{p_{int}^{ik}\}$. 

\subsubsection{Attention Mask}
To calculate the attention mask of all persons in the image, we first obtain the body-part boxes from the pose estimation~\cite{fang2017rmpe,li2019crowdpose,li2021human} results following~\cite{li2021transferable}. 
For the failure cases of pose estimation, we use the whole-body detection boxes from~\cite{ren2015faster} as body-part boxes.
With the $k$-th body-part boxes as $b(p)^k=\{b(p)^{kl} | b(p)^{kl} = [w^{kl}_1, h^{kl}_1, w^{kl}_2, h^{kl}_2], 1 \le l \le L\}$ ($L$ is the persons count in the image),
the global body-parts mask matrix $m^{ik}_{part}$ 
($\forall i , m^{ik}_{part}=m^{k}_{part}$) 
is calculated as:
\begin{equation}
m^{k(xy)}_{part} = \left\{
            \begin{array}{rcl}
1       &      & {\exists ~l, h^{kl}_1 \le  x*(H_0/H) \le h^{kl}_2,  w^{kl}_1 \le  y*(W_0/W) \le w^{kl}_2}\\
0       &      & {otherwise}
\end{array} \right.
,
\end{equation}
where $x, y$ is the index of the matrix. The scaling factor $H_0/H$ and $W_0/W$ are used here because the size of the feature map $z$ is scaled down from that of the original image, 
\eg, $H_0/H=W_0/W=32$ with ResNet-50 backbone. 

Notably, although one proposal corresponds to only \textit{one} targeted H-O pair, its body-part saliency map contains the body-parts of \textit{all} persons in the image. Thus, the relationship between body-parts of all persons is learned from a global perspective.
Here $m^{ik}_{part}$  is a core component of the final mask $m^{ik}$. 
We briefly set $m^{ik}=m^{ik}_{part}$ for the intuitive scheme in this section, and will further modify $m^{ik}$ in the next sections.

\subsection{Progressively Body-Part Masking for Flexibility}
\label{sec:progressive}
In Sec.~\ref{sec:intuitive}, an intuitive scheme with body-part saliency map is introduced, and its effectiveness is validated by experiments (Sec.~\ref{sec:exp}).
However, it lacks flexibility to \textit{simply} construct body-part saliency maps to highlight all the \textbf{same} body-parts.
As aforementioned in Fig.~\ref{fig:intro}b, it also matters to learn attention from \textbf{different} body-parts of \textbf{different} persons.
Thus, we develop a \textit{progressively masking} strategy, where different attention masks are applied in successive transformer layers.
Considering that the token representations are encoded more and more sufficiently throughout the whole network, fewer tokens are dropped in the early layers while more tokens are dropped in the late layers. In other words, the ``\textbf{receptive field}'' gradually becomes more focused as the computation proceeds. Attention to different body-parts of different persons is learned in the early layer, then comes the same body-parts of different persons. After encoding useful information from a global perspective, our model focuses on the targeted person in the final layer, which is similar to TIN++~\cite{li2021transferable}.

\subsubsection{Implementation}
We construct the progressive masks from three components: body-part mask $m_{part}^k$ ($1 \le k \le 6$, Sec.~\ref{sec:intuitive}) shared by proposals, detected human mask $m_{hum}^i$ ($1 \le i \le N_q$) dynamically calculated from the box detection of the $i$-th proposal, and detected object mask $m_{obj}^i$ ($1 \le i \le N_q$) similar to $m_{hum}^i$.
For the $i$-th proposal and its detected human bounding box $b(h)^i \in [0, 1]^{4}$, the box is first unnormalized by the image size as $b(h)^i=[w^{i}_1, h^{i}_1, w^{i}_2, h^{i}_2]$. Then 
\begin{equation}
m_{hum}^{i(xy)} = \left\{
            \begin{array}{rcl}
1       &      & {h^{i}_1 \le  x*(H_0/H) \le h^{i}_2,  w^{i}_1 \le  y*(W_0/W) \le w^{i}_2}\\
0       &      & {otherwise}
\end{array} \right.
,
\end{equation}
The detected object mask $m_{obj}^i$ is obtained from $b(o)^i$ in a similar way.

The transformer decoder $f_{dec2}$ for interactiveness inference has three stacked layers.
The attention mask $m^{ik}_{j} \in \{0,1\}^{H \times W}$ for the $i$-th proposal, the $j$-th ($j=1,2,3$, larger for later layer) layer and the $k$-th body-part is:
\begin{align}
m^{ik}_1 & = max(m_{hum'}^i, m_{part}^k, m_{obj}^i), m_{hum'}^i= max(\mathop{max}\limits_k(m_{part}^k)-m_{hum}^i), 0),\\
m^{ik}_2 &= max(m_{part}^k, m_{obj}^i), \\
m^{ik}_3 &= max(min(m_{part}^k, m_{hum}^i), m_{obj}^i).
\end{align}
An example is given in Fig.~\ref{fig:module}b.
The targeted object is highlighted in all layers.
Specifically, the 1-st layer highlights the $k$-th body-part of the detected human, and the whole body of the other humans in the image ($m_{hum'}^i$), which allows attention computation from different body-parts of different persons. 
The 2-nd layer emphasizes the $k$-th body-part from all persons in the image, while the 3-rd layer focuses on the $k$-th body-part of the targeted person.
With the progressive mask throughout transformer layers, different visual patterns are flexibly encoded to facilitate body-part interactiveness learning. 

\subsection{One-Time Passing via Body-Parts Filtering and Merging} 
\label{sec:filter&merge}
In the intuitive scheme, six repeated calculations are needed for each body-part because their saliency maps are not shared. However, this is computationally redundant since body-part interactiveness has a notable property: \textbf{sparsity}. Namely, only several body-parts will get involved when people are interacting with daily objects. For instance, in ``eat apple'', the head and hands have stronger relationships with the apple than the lower body.
Therefore, we can select the most important body-parts and only feed their visual feature into the network to classify interactiveness, \ie, the \textit{filtering} process. 
Then, for the selected body-parts, there are still repeated calculations although fewer times. Thus, we spatially \textit{merge} the saliency maps of the selected body-parts, feed them into the network, and obtain the whole-body interactiveness prediction via a one-time calculation. The rationality of merging is validated from the following two aspects. First, it seems that the spatial merging leads to ``mixed-up'' body-part feature, since merging all the six body-parts is equal to instance-level learning instead of body-part level. However, we emphasize that body-part level finer-grained feature remains in our model because only \textit{several} body-parts are merged. Second, here the whole-body interactiveness prediction is directly calculated, while in the intuitive scheme we obtain it from six body-part interactiveness predictions. This does not impede interactiveness learning because most \textit{important} body-parts are used for calculation.

\subsubsection{Implementation} 
The implementation is illustrated in Fig.~\ref{fig:module}b. 
First, the importance of body-parts is calculated via a one-layer transformer decoder layer $h(\cdot,\cdot, \cdot)$ without mask. 
For the $i$-th proposal, it transforms $d^i$ into $d^{i}_{part} = h(z, d^i, pos)$, and a subsequent FFN is applied to get body-part attention score $\{p^{ik}_{part}\}_{k=1}^6=Sigmoid(FFN(d^{i}_{part}))$. Then, body-parts with relatively higher scores (top 25\% of $\{p^{ik}_{part} | 1 \le i \le N_q, 1 \le k \le 6\}$) are chosen for the following computation and others are filtered out. The result is represented as $\{n^{ik} \in \{0, 1\}\}$ (1 for chosen and 0 for dropped).
After the filtering, 
both the mask matrix and query embeddings are merged as:
\begin{align}
    m^i_{j} &=max_k(m^{ik}_j * n^{ik}),\\
    d^{i}_{mer} &= d^{i} + \sum_k(d_{part}^{ik} * n^{ik} * p^{ik}_{part}).
\end{align}
For the mask matrix, the spatial merge is equivalent to the elementwise maximum operation. 
For the query embeddings, the weighted sum of query embeddings $d_{part}^{ik}$ of the selected body-parts is added to the whole-image query embeddings $d^{i}$ to generate the merged embeddings $d^i_{mer}$. Then the decoder $f_{dec2}$ with progressively masking transforms $d^i_{mer}$ into embedding $e^i_{mer}$.
Finally, the interactiveness score is obtained by a FFN via $p^i_{int}=Sigmoid(FFN(e^{i}_{mer}))$.
Compared with the intuitive scheme, the one-time passing strategy drops unnecessary computation and improves computation efficiency. 

\subsection{Training and Inference}
When training our model, following the set-based training process of QPIC~\cite{qpic}, we first match predictions and ground-truths via bipartite matching, then calculate the loss for the matched pairs.
For box detection, the loss is composed by three parts: box regression loss $L_b$, intersection-over-union loss $L_u$ and object-class loss $L_c$. The loss $L_{det}$ is obtained via $L_{det}=\lambda_1 * L_b + \lambda_2 * L_u + \lambda_3 * L_c$, where $\lambda_1, \lambda_2, \lambda_3$ are hyperparameters for adjusting the weights.
The interactiveness classifier and verb classifier are supervised with classification loss $L_{int}$ and $L_{verb}$ respectively.

The training is divided into two stages. First, we train the box detector and interactiveness classifier along with the visual feature extractor with loss $L_1=L_{det}+L_{int}$. Then, the box detector and verb classifier are trained along with the visual feature extractor and the loss is $L_2=L_{det}+L_{verb}$. 
In inference, we use the interactiveness results to improve verb classification via non-interaction suppression (NIS)~\cite{li2021transferable}, where H-O pairs with lower interactiveness scores are filtered out.

\section{Discussion: Sparse vs. Crowded Scene}

In this section, we discuss a naturally raised question: our method focuses on crowded scenes with multi interactive pairs or multi persons/objects, then what about the sparse scenes?

First, our model is adapted to both crowded and sparse scenes. Under crowded scenes, mining cross-person cues provides more useful information of interactiveness. While for sparse scenes, our method would be operated similar to TIN++~\cite{li2021transferable}. 
However, our model is still superior to TIN++~\cite{li2021transferable} thanks to the integration body-part feature and DETR~\cite{detr} backbone.

Next, we want to re-emphasize the importance of HOI detection in crowded scenes, especially for interactiveness learning and interactive human-object pair proposal detection.
We find that crowded images occupy a large proportion of the HOI dataset, validating the effects brought by our method. 
Meanwhile, the performance of interactiveness learning under crowded scenes is inferior to sparse ones. We split HICO-DET~\cite{hicodet} test set into \textit{sparse/crowded} scenes respectively and evaluate the interactiveness AP: 
16.96/9.64 AP (TIN++~\cite{li2021transferable}) and  43.62/33.10 AP (ours, Sec.~\ref{sec:exp}). 
From the large performance gap between sparse and crowded scenes (7.32 AP for TIN++~\cite{li2021transferable} and 10.52 AP for ours, Sec.~\ref{sec:exp}), 
we can see that interactiveness learning is mainly bottlenecked by crowded scenes, where it is usually harder to effectively extract interactive pairs. Therefore, it matters to focus on crowded scenes for interactiveness learning. Statistics are detailed in the supplementary.

Then, we further propose a novel \textbf{sparsity adaptive sampling strategy} on train set to put more emphasis on crowded scenes and facilitate interactiveness learning. The sampling probability is modified as 1:$\alpha$ ($\alpha$ \textgreater 1, in practice $\alpha=3$) for sparse vs. crowded images, compared with the original 1:1,
which guides better interactiveness knowledge to identify interactive pairs under complex scenes.
Finally, the experiment (Sec.~\ref{sec:result}) proves that the proposed global perspective indeed boosts interactiveness learning, especially for crowded scenes.

\section{Experiment}\label{sec:exp}

In this section, we first introduce the datasets and metrics in Sec.~\ref{sec:dataset}, and describe implementation details in Sec.~\ref{sec:imple}. Next, we report the results on HICO-DET~\cite{hicodet} and V-COCO~\cite{vcoco} in Sec.~\ref{sec:result}. Some visualization results are given in Sec.~\ref{sec:vis}. Finally, ablation studies are conducted in Sec.~\ref{sec:ablation}.

\subsection{Dataset and Metric}\label{sec:dataset}

\noindent{\bf Datasets.} We adopt two datasets HICO-DET~\cite{hicodet} and V-COCO~\cite{vcoco}. HICO-DET~\cite{hicodet} includes 47,776 images (38,118 in train set and 9658 in test set), 600 HOI categories on 80 object categories (same with ~\cite{coco}) and 117 verbs, and provides more than 150k annotated human-object pairs. V-COCO~\cite{vcoco} provides 10,346 images (2,533 for training, 2,867 for validating, and 4,946 for testing) and 16,199 person instances. Each person has labels for 29 action categories (five of them have no paired object). 

\noindent{\bf Metrics.} We follow the settings adopted in~\cite{hicodet}, i.e., a prediction is a true positive only when the human and object bounding boxes both have IoUs larger than 0.5 with reference to ground truth, and the HOI classification result is accurate. The role mean average precision~\cite{vcoco} is used to measure the performance. Additionally, we measure interactiveness detection in a similar setting. 

\subsection{Implementation Details}\label{sec:imple}
We adopt ResNet-50 followed by a six-layer transformer encoder as our visual feature extractor. The box detector and the verb classifier are both implemented as a six-layer transformer decoder. The interactiveness classifier is implemented as a three-layer transformer decoder, where selected tokens are masked. 
During training, AdamW~\cite{adamw} with the weight decay of 1e-4 is used. The visual feature extractor and box decoder are initialized from COCO~\cite{coco} pretrained DETR~\cite{detr}. The query size is set as 64 for HICO-DET~\cite{hicodet} and 100 for V-COCO~\cite{vcoco} following CDN~\cite{cdn}. The loss weight coefficients $\lambda_1, \lambda_2, \lambda_3$ are respectively set as 1, 2.5, 1, exactly the same as QPIC~\cite{qpic}. 
In 1st stage, the model is trained for 90 epochs with a learning rate of 1e-4 which is decreased by 10 times at the 60th epoch. In 2nd stage, the model is fine-tuned for 60 epochs.
All experiments are conducted on four NVIDIA GeForce RTX 3090 GPUs with a batch size of 16. 
In inference, a pairwise NMS with a threshold of 0.6 is conducted. 

\subsection{Results}\label{sec:result}
\subsubsection{Interactiveness Detection.}
We evaluate our interactiveness detection on HICO-DET~\cite{hicodet} and V-COCO~\cite{vcoco}. On HICO-DET~\cite{hicodet}, we adopt the interactiveness detection AP proposed in TIN++~\cite{li2021transferable}, while on V-COCO~\cite{vcoco} we construct the benchmark in a similar way.
Tab.~\ref{tab:binary} shows our interactiveness detection results compared with open-source state-of-the-art methods~\cite{li2021transferable,ppdm,qpic,cdn}. For TIN++~\cite{li2021transferable}, the output interactiveness score is used. For PPDM~\cite{ppdm}, QPIC~\cite{qpic}, and CDN~\cite{cdn}, the mean of HOI scores (520 HOI categories for HICO-DET~\cite{hicodet}, and 24 for V-COCO~\cite{vcoco}) is used as an approximation.

\begin{table}
\centering
\resizebox{0.45\textwidth}{!}{
\begin{tabular}{l c c}
\hline
Method   & HICO-DET~\cite{hicodet} & V-COCO~\cite{vcoco}\\
\hline
TIN++~\cite{li2021transferable} & 14.35 & 29.36\\
PPDM~\cite{ppdm}  & 27.34 & - \\
QPIC~\cite{qpic} & 32.96 & 38.33\\
CDN~\cite{cdn}  & 33.55 & 40.13\\
\hline
ours & \textbf{38.74} (+\textbf{5.19}) & \textbf{43.61} (+\textbf{3.48})\\
\hline
\end{tabular}}
\caption{Interactiveness detection results on HICO-DET~\cite{hicodet} and V-COCO~\cite{vcoco}.}
\label{tab:binary}
\end{table}

As is shown in Tab.~\ref{tab:binary}, for the two-stage method TIN++~\cite{li2021transferable}, the interactiveness AP is unsatisfactory. We claim that it suffers from exhaustively generated negative H-O pairs from the detector in the first stage, despite the non-interaction suppression~\cite{li2021transferable}. 
Instead, the one-stage methods PPDM~\cite{ppdm}, QPIC~\cite{qpic}, and CDN~\cite{cdn} benefit from the avoidance of exhaustive pairing and achieve better performance on interactiveness detection. 
With the insight of holistically modeling body-part level interactiveness, our method achieves a even better interactiveness AP of 
\textbf{38.74} on HICO-DET~\cite{hicodet} and \textbf{43.61} on V-COCO~\cite{vcoco}.

Next, we evaluate the effectiveness of our method in hard cases, as is discussed in Sec.~\ref{sec:intro}. We split HICO-DET~\cite{hicodet} test set and compare the interactiveness detection performance of TIN++~\cite{li2021transferable} and our method.
Compared with the local-based method TIN++~\cite{li2021transferable}, 
our method improves an interactiveness AP gain of 26.66 (157\%) / 23.46 (243\%) for sparse/crowded scenes,  23.74 (147\%) / 23.53(263\%) for normal/tiny-persons scenes, and 22.11 (134\%) / 14.69 (182\%) for less/more-occlusion scenes. We can see that the proposed global perspective indeed boosts interactiveness learning, especially for hard cases where interactive persons are more difficult to identify.
For detailed results, please refer to our supplementary. 

\begin{table}
\centering
\resizebox{0.6\textwidth}{!}{
\begin{tabular}{l  c  c  c  c  c  c  }
\hline
         & \multicolumn{3}{c}{Default}  &\multicolumn{3}{c}{Known Object} \\
Method         & Full & Rare & Non-Rare  & Full & Rare & Non-Rare \\
\hline
\hline
iCAN~\cite{gao2018ican}   & 14.84  & 10.45  & 16.15  & 16.26  & 11.33  & 17.73\\
TIN~\cite{li2019transferable}  & 17.03  & 13.42  & 18.11  & 19.17  & 15.51  & 20.26\\
PMFNet~\cite{pmfnet}  &  17.46  & 15.65  & 18.00  & 20.34  & 17.47  & 21.20\\
DJ-RN~\cite{djrn}    & 21.34  & 18.53 &  22.18 &  23.69 &  20.64  & 24.60\\
\hline
PPDM~\cite{ppdm}  & 21.73 & 13.78 & 24.10 & 24.58 &  16.65  & 26.84 \\
VCL~\cite{vcl}   &  23.63  & 17.21 &  25.55 &  25.98 &  19.12 &  28.03 \\
IDN~\cite{li2020hoi}  & 26.29  & 22.61 &  27.39  & 28.24  & 24.47  & 29.37\\
Zou \etal~\cite{zou2021end}  & 26.61  & 19.15  & 28.84 &  29.13  & 20.98  & 31.57\\
AS-Net~\cite{AS-Net}  & 28.87  & 24.25 &  30.25 &  31.74  & 27.07  & 33.14\\
QPIC~\cite{qpic}  & 29.07 &  21.85 &  31.23 &  31.68  & 24.14 &  33.93\\
FCL~\cite{hou2021fcl}  & 29.12  & 23.67 &  30.75  & 31.31  & 25.62  & 33.02\\
GGNet~\cite{GGNet}  & 29.17  & 22.13  & 30.84  & 33.50  & 26.67  & 34.89\\
CDN~\cite{cdn}  & 31.78  & 27.55  & 33.05  & 34.53  & 29.73  & 35.96\\
\hline
Ours  & \textbf{35.15}  & \textbf{33.71} &  \textbf{35.58}  & \textbf{37.56}   &  \textbf{35.87}  & \textbf{38.06}\\
\hline
\end{tabular}}
\caption{Results on HICO-DET~\cite{hicodet}.}
\label{tab:hico-det}
\end{table}

\subsubsection{Body-Part Interactiveness Detection.}
To detail the analysis of body-part level interactiveness learning, we evaluate body-part interactiveness AP from the output body-part attention score. 
When trained without body-part interactiveness supervision, our method still learns body-part interactiveness well. The interactiveness APs on HICO-DET~\cite{hicodet} are:  38.74 (whole body), 11.66 (feet), 5.31 (legs) , 23.83 (hip), 23.11 (hands), 1.38 (arms), 5.51 (head).
Further, we utilize annotations provided by HAKE~\cite{hake,li2022hake} to apply body-part supervision,
\textit{i.e.}, $\{p^{ik}_{part}\}_{k=1}^6$ is bound with body-part level interactiveness labels, and the loss is added to $L_{int}$.
The results are: 38.83 (whole body), 34.89 (feet), 31.02 (legs) , 38.11 (hip), 34.18 (hands), 31.32 (arms), 24.94 (head).
We can see that the performance is further improved with body-part supervision, especially for ``legs'', ``arms'', and ``head''.
Without supervision, the performance of ``arms'' and ``head'' are inferior to other body-parts. 
One possible reason is that ``arms'' suffer from relatively ambiguous definitions, and can sometimes be confused with ``torso'' or ``hands'' due to occlusion. Additionally, ``head'' is related to HOIs harder to identify such as ``look'', ``smell''.

\begin{table}
\centering
\resizebox{0.4\textwidth}{!}{
\begin{tabular}{l  c  c }
  \hline
Method         &$AP_{role}$(S1) & $AP_{role}$(S2) \\
\hline
\hline
iCAN~\cite{gao2018ican}  & 45.3  & 52.4\\
TIN~\cite{li2019transferable}  & 47.8  & 54.2\\
VSGNet~\cite{vsgnet}  & 51.8  & 57.0\\
PMFNet~\cite{pmfnet}  & 52.0  & -\\
IDN~\cite{li2020hoi}  & 53.3  & 60.3\\
AS-Net~\cite{AS-Net} & 53.9  & -\\
GGNet~\cite{GGNet} & 54.7 &  -\\
HOTR~\cite{HOTR}  & 55.2  & 64.4\\
QPIC~\cite{qpic}  & 58.8  & 61.0\\
CDN~\cite{cdn}  & 62.3  & 64.4\\
\hline
Ours   & \textbf{63.0} & \textbf{65.1}\\
\hline
\end{tabular}}
\caption{Results on VCOCO~\cite{vcoco}.}
\label{tab:vcoco}
\end{table}

\subsubsection{HOI Detection Boosting.}
In Tab.~\ref{tab:hico-det} and Tab.~\ref{tab:vcoco}, we evaluate how HOI learning can benefit from the interactiveness detection results of our method. 
Here we use instance-level supervision without annotations from HAKE~\cite{hake,li2022hake} for interactiveness learning to compare with TIN++~\cite{li2021transferable}.
In Tab.~\ref{tab:hico-det}, the first part adopted COCO pre-trained detector. HICO-DET fine-tuned or one-stage detector is used in the second part. All the results are with ResNet-50.

Our method outperforms state-of-the-arts with \textbf{35.15}/\textbf{37.56} mAP (Default Full/ Known Object Full) on HICO-DET~\cite{hicodet} and \textbf{63.0}/\textbf{65.1} mAP (Scenario 1/2) on V-COCO~\cite{vcoco}, verifying its effectiveness.
For two-stage HOI methods, we feed the representative method iCAN~\cite{gao2018ican} (human-object pairs are exhaustively paired) with our detected pairs. 
Tab.~\ref{tab:two-stage} reports the performance comparison on HICO-DET~\cite{hicodet} with different pair detection results.
We find that with high-quality detected interactive pairs, the performance of iCAN~\cite{gao2018ican} is significantly boosted, especially from the results of our method. We leave the detailed settings in supplementary.

\begin{table}
\centering
\resizebox{0.4\textwidth}{!}{
\begin{tabular}{lccc}
    \hline
Method         &Full & Rare & Non-Rare \\
\hline
$iCAN$ & 14.84  & 10.45  & 16.15 \\
${iCAN}^{QPIC}$  & 20.36 & 11.14 & 23.11 \\
${iCAN}^{CDN}$  & 21.09 & 11.20 & 24.04\\
${iCAN}^{Ours}$  & \textbf{24.38} & \textbf{16.27} & \textbf{26.80}\\
\hline
\end{tabular}}
\caption{The performance comparison on HICO-DET~\cite{hicodet} with different pair detection results.}
\label{tab:two-stage}
\end{table}

\begin{table}
\centering
\resizebox{0.6\textwidth}{!}{
\begin{tabular}{lcc}
   \hline
Method         & int AP& HOI mAP\\
\hline
\hline
Ours & \textbf{38.74} & \textbf{35.15} \\
\hline 
w/o body-part & 36.46 & 32.16\\ %
w/o body-part saliency map & 37.43 & 32.60 \\ %
intuitive scheme & 37.91 & 33.12\\ %
\hline
w/o progressive mask & 38.06 & 34.05\\ %
w/o sparsity adaptive sampling & 38.29 & 34.37\\ %
w/o one-time passing & 38.51&34.90\\ %
\hline
\end{tabular}}
\caption{Results of ablation studies on HICO-DET~\cite{hicodet}.}
\label{tab:ablation}
\end{table}

\subsection{Visualization}\label{sec:vis}

\begin{figure}
\centering
\includegraphics[width=0.8\textwidth]{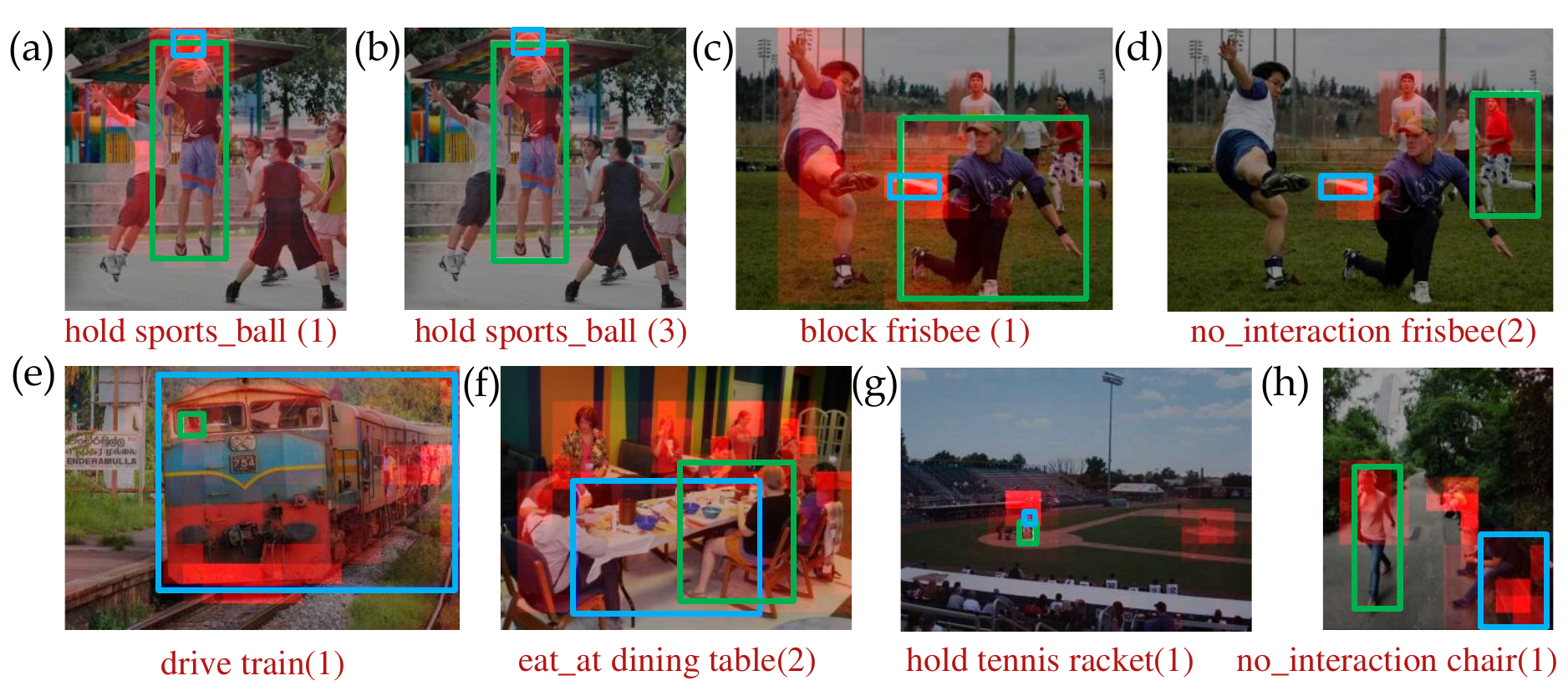}
\caption{Visualization results of learned attention. Number $j \ (j=1,2,3)$ in the bracket represents which layer the attention results are obtained from.}
\label{fig:vis}
\end{figure}

Fig.~\ref{fig:vis} shows some visualization results of the learned attention. 
Our model can effectively learn body-part attention (Fig.~\ref{fig:vis}b) and extract informative cues from other persons in the image, either from the same (Fig.~\ref{fig:vis}d, f) or the different (Fig.~\ref{fig:vis}a, c) body-parts. Learning holistic relationship between body-parts from different persons alleviates the difficulties of interactiveness learning in hard cases, \eg, tiny interactive persons (Fig.~\ref{fig:vis}e, g), crowded scenes (Fig.~\ref{fig:vis}e), and occlusion (Fig.~\ref{fig:vis}f).
Also, our method benefits both interactive pairs and non-interactive pairs (Fig.~\ref{fig:vis}d, h). 
For more please refer to our supplementary.

\subsection{Ablation Studies}\label{sec:ablation}
We conduct ablation studies on HICO-DET~\cite{hicodet} and list interactiveness AP and corresponding HOI mAP results in Tab.~\ref{tab:ablation}.

We first validate the effectiveness of the body-part saliency map. In the intuitive scheme, the body-part saliency map is applied via an attention mask and interactiveness is calculated for each body-part. The intuitive scheme achieves 37.91 interactiveness AP and 33.12 HOI mAP. 
In contrast, the interactiveness AP falls to 37.43 when body-part saliency map is removed, 
\textit{i.e.}, for each proposal, body-part and layer, the mask matrix is set as $\mathbf{1}^{H \times W}$. 
It validates the effectiveness of introducing global-level visual cues and learning holistic body-part relationship. 
Further, if trained with only instance-level visual features without emphasis on body-parts, the interactiveness AP falls to 36.46. Thus, learning body-part interactiveness benefits instance-level interactiveness learning by introducing fine-grained visual features.

Then we evaluate the proposed modules. 
(1) When progressively masking is removed, all decoder layers are applied with the same attention mask of body-part saliency map, which leads to an interactiveness AP drop to 38.06. 
The result validates that our model indeed benefits from the progressively masking strategy, where diverse body-part oriented visual patterns are combined and integrated to facilitate interactiveness learning.
(2) Without sparsity adaptive sampling, the interactiveness AP falls to 38.29. We find that the sampling augmentation on crowded-scene images helps to extract interactive pairs,  especially in complex scenes.
(3) Finally, we evaluate the one-time passing strategy. If removed, the interactiveness AP falls to 38.51, and computation speed is reduced by approximately 20\% from 28 min/epoch to 33 min/epoch. We can see that our model benefits from it to improve performance as well as efficiency.

\section{Conclusions}
Currently, HOI detection is still bottlenecked by interactiveness learning. In this paper, we focus on learning human body-part interactiveness from a previously overlooked global perspective. We construct body-part saliency maps to mine informative cues from not only the targeted person, but also other persons in the image. Our method provides an effective supplement for existing local-based methods and achieves significant improvements on widely used HOI benchmarks. 
Despite the improvement, we believe there is still much room left to make further progress on interactiveness learning.

\vspace{0.3cm}
\noindent{\textbf{Acknowledgements.}}
This work was supported by the National Key R\&D Program of China (No. 2021ZD0110700), Shanghai Municipal Science and Technology Major Project (2021SHZDZX0102), Shanghai Qi Zhi Institute, and SHEITC (2018-RGZN-02046).

\clearpage
\section*{Supplementary Materials}
\appendix
\section{HOI Hard Cases}
In Fig.~\ref{fig:intro}a, we show the histograms of hard cases in HICO-DET~\cite{hicodet} train set and find that hard cases are common in HOI datasets, which impedes interactiveness learning. 
In Sec.~5.3, we split HICO-DET~\cite{hicodet} test set and compare the interactiveness detection performance of TIN++~\cite{li2021transferable} and our method. 
We detail the settings as follows:
\begin{itemize}
    \item \textbf{Tiny interactive persons.} 
    In the 1-st histogram in Fig.~1a, for each interactive H-O pair, the ratio $r$ of the area of the human bounding box to the area of the image is calculated. In the test set, the image is considered as ``tiny-persons scenes'', if it has at least one interactive pair with the ratio $r \textless 0.1$.
    \item \textbf{Crowded Scenes.}
    We calculate the detected person counts in each image and show them in the 2-nd histogram in Fig.~1a. In the test set, the image is considered as ``crowded scenes'', if it has at least three interactive pairs.
    \item \textbf{Occlusion.}
    In the 3-rd histogram in Fig.~1a, for each interactive H-O pair, the average of human joint detection confidence is calculated based on the pose estimation results~\cite{fang2017rmpe}. In the test set, for each image, we calculate the average of joint detection confidence of all the human bounding boxes in it as $j$. An image is regarded as ``more-occlusion scenes'' if $j \textless 0.2$ and regarded as ``less-occlusion scenes'' if $j \textgreater 0.6$.
\end{itemize}

Notably, we split the test set in \textit{image level} for the convenience of model inference, 
\ie, avoid an image to be seen in 
inference phrase.
For example, in an image tiny persons and normal size persons may co-exist, and more/less-occluded persons may co-exist. 
After the split, 
for all the interactive persons in ``tiny-persons scenes'' images, 84.3\% of them has a ratio $r \textless 0.1$. 
For all the interactive persons in ``more-occlusion scenes'' images, 83.1\% of them has an average of joint detection confidence $j \textless 0.2$. 
Thus, with the image-level split, the model performance on hard cases can still be evaluated effectively and accurately with a much higher ratio of hard instances within images compared to the previous split.

Tab.~\ref{tab:hard-cases} shows interactiveness AP under different settings on HICO-DET~\cite{hicodet}.
With our global perspective to learn the holistic relationship of body-part interactiveness, the difficulty of interactiveness learning is alleviated. Comparing our method with open-source state-of-the-art methods~\cite{li2021transferable,ppdm,qpic,cdn}, the gains of hard cases are larger than non-hard cases, validating the effectiveness of our method, especially for HOI hard cases.

\begin{table}
\centering
\caption{Interactiveness AP under different settings.}
\label{tab:hard-cases}
\resizebox{0.7\textwidth}{!}{
\begin{tabular}{l c c c c}
\hline
Method   & Full & Sparse/Crowded & Normal/Tiny & Less/More Occ\\
\hline
TIN++ & 14.35 & 16.96/9.64 & 16.11/8.94 & 16.49/8.06\\
PPDM & 27.34 & 34.67/26.69 & 31.79/26.33 & 29.83/17.25\\
QPIC & 32.96 & 36.80/27.04 & 34.02/26.14 & 32.08/19.75\\
CDN & 33.55 & 39.92/28.84 & 36.10/25.11 & 34.55/21.69\\
\hline
Ours & \textbf{38.74} & \textbf{43.62}/\textbf{33.10} & \textbf{39.85}/\textbf{32.47} & \textbf{38.60}/\textbf{22.75}\\
\hline
\end{tabular}}
\end{table}

\section{Sparse vs. Crowded Scene}
In this section, we detail the discussion about sparse/crowded scenes in Sec.~4 of the main text.

In the widely used HOI dataset HICO-DET~\cite{hicodet} and V-COCO~\cite{vcoco}, we analyze the detected person counts on each image, and find that images with more than two persons account for 47.3\%/62.5\% in train/test set in HICO-DET~\cite{hicodet}, and the number is 40.36\%/58.6\% for V-COCO~\cite{vcoco}. Thus, crowded images occupy a large proportion in the HOI dataset, validating the effects brought by our method. 

We split HICO-DET~\cite{hicodet} test set into \textit{sparse} and \textit{crowded} scenes respectively and evaluate the interactiveness AP. Here, images with at least three interactive pairs are considered as ``crowded''. The performances are 16.96/9.64 AP (TIN++~\cite{li2021transferable}) and  43.62/33.10 AP (ours). From the large performance gap (7.32 for TIN++~\cite{li2021transferable} and 10.52 for ours), we can see that interactiveness learning is mainly bottlenecked by crowded scenes. Therefore, it matters to focus on crowded scenes for interactiveness learning.

\section{Detailed Experiment Settings}
In our training process, the interactiveness classifier is first trained and then is the verb classifier. Since the box detector is fine-tuned in the 2-nd stage, for each image we finally get two predicted sets: 1) detection with verb classification results $p_v$: $R = \{r^i | r^i = (b(h)^i, b(o)^i, c^i, p_{verb}^i)\}_{i=1}^{N_q}$, and 2) detection with interactiveness classification results $p$ : $R^{'} = \{r^{j'} | r^{j'} = (b(h)^{j'}, b(o)^{j'}, c^{j'}, p^j_{int})\}_{j=1}^{N_q}$. We use $R^{'}$ to calculate interactiveness AP in Tab.~1. For HOI detection boosting in Tab.~2 and Tab.~3, for each $r^i \in R$, a matched proposal $r^{f(i)'}$ is found from $R^{'}$ and the matched interactiveness results $p^{f(i)}_{int}$ is used for non-interaction suppression~\cite{li2019transferable}, where H-O pairs with lower interactiveness scores $p^{f(i)}_{int}$ are filtered out. Here, the matching function $f(i)$ is obtained via
\begin{equation}
f(i) = \mathop{argmax}\limits_{j: 1 \le j \le N_q, c^i = c^{j'}} (IoU(b(h)^i, b(h)^{j'}) + IoU(b(o)^i, b(o)^{j'})).
\end{equation}
\ie, the matched proposal should have the correct object class and maximum human\&object bounding boxes IoUs. When no matched proposal is found for $r^i$, the interactiveness score is 0. 

In Tab.~4, we feed the representative two-stage HOI method iCAN~\cite{gao2018ican} (human-object pairs are exhaustively paired) with our detected pairs. 
The inference score $S$ is obtained via $S = S_v * S_o$, where $S_v$ is the HOI prediction score from iCAN~\cite{gao2018ican} and $S_o$ is the object prediction score from our method. Also, a pairwise NMS with a threshold of 0.6 is conducted. In our method, interactiveness inference results are used for filtering out non-interaction pairs.
Additionally, CDN~\cite{cdn} reports similar results by replacing detected boxes (Tab.~1 in their paper) while the results are different. We assume it is because of different experiment settings.

\begin{figure}
\centering
\includegraphics[width=\textwidth]{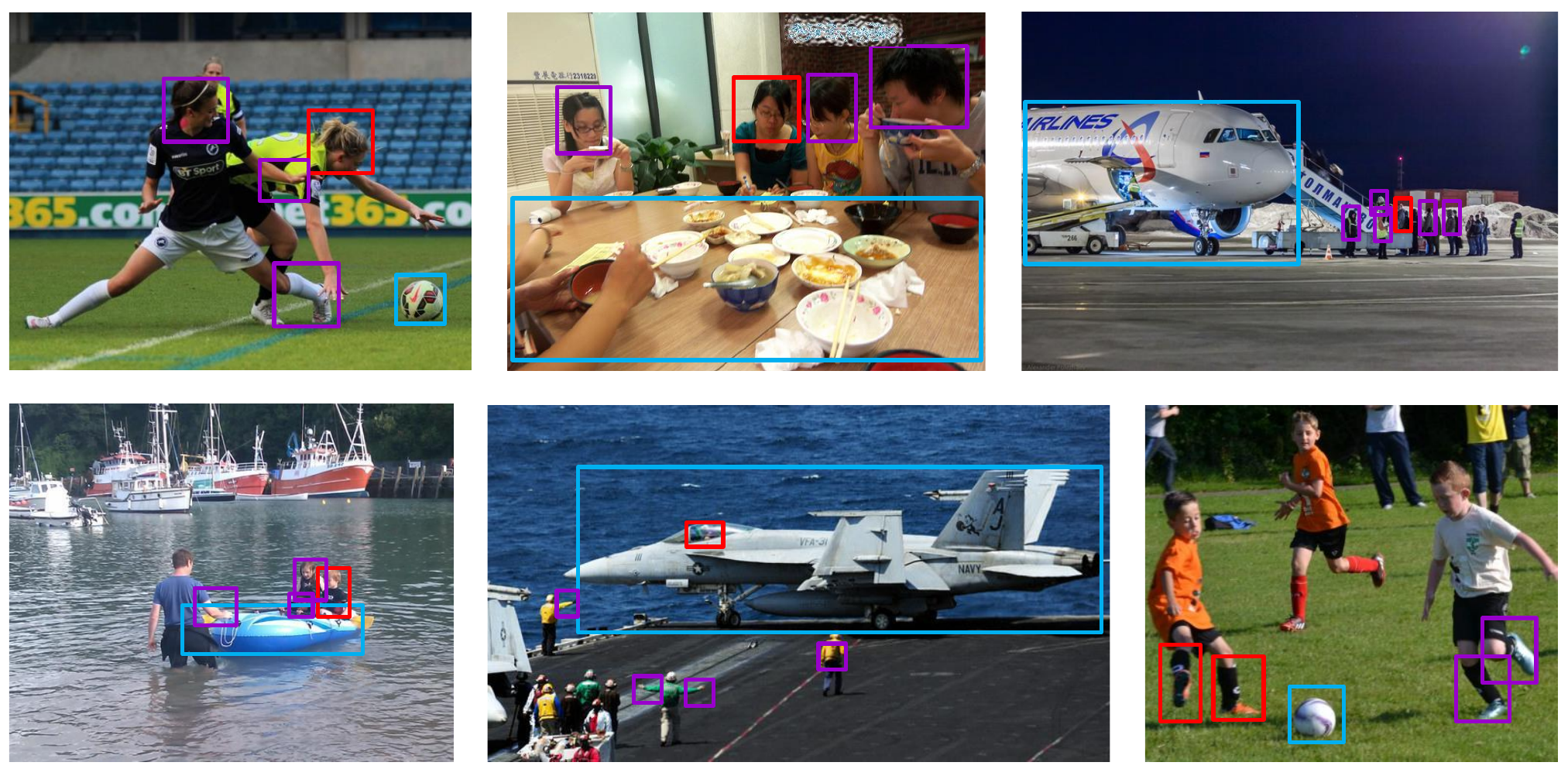}
\caption{Some examples where our proposed global perspective facilitates interactiveness learning. In the images, red boxes represent targeted body-parts, blue boxes represent targeted objects, and purple boxes represent body-parts of other persons which provide informative cues for interactiveness classification of the targeted body-parts and objects.}
\label{fig:examples}
\end{figure}

\begin{figure}
\centering
\includegraphics[width=0.8\textwidth]{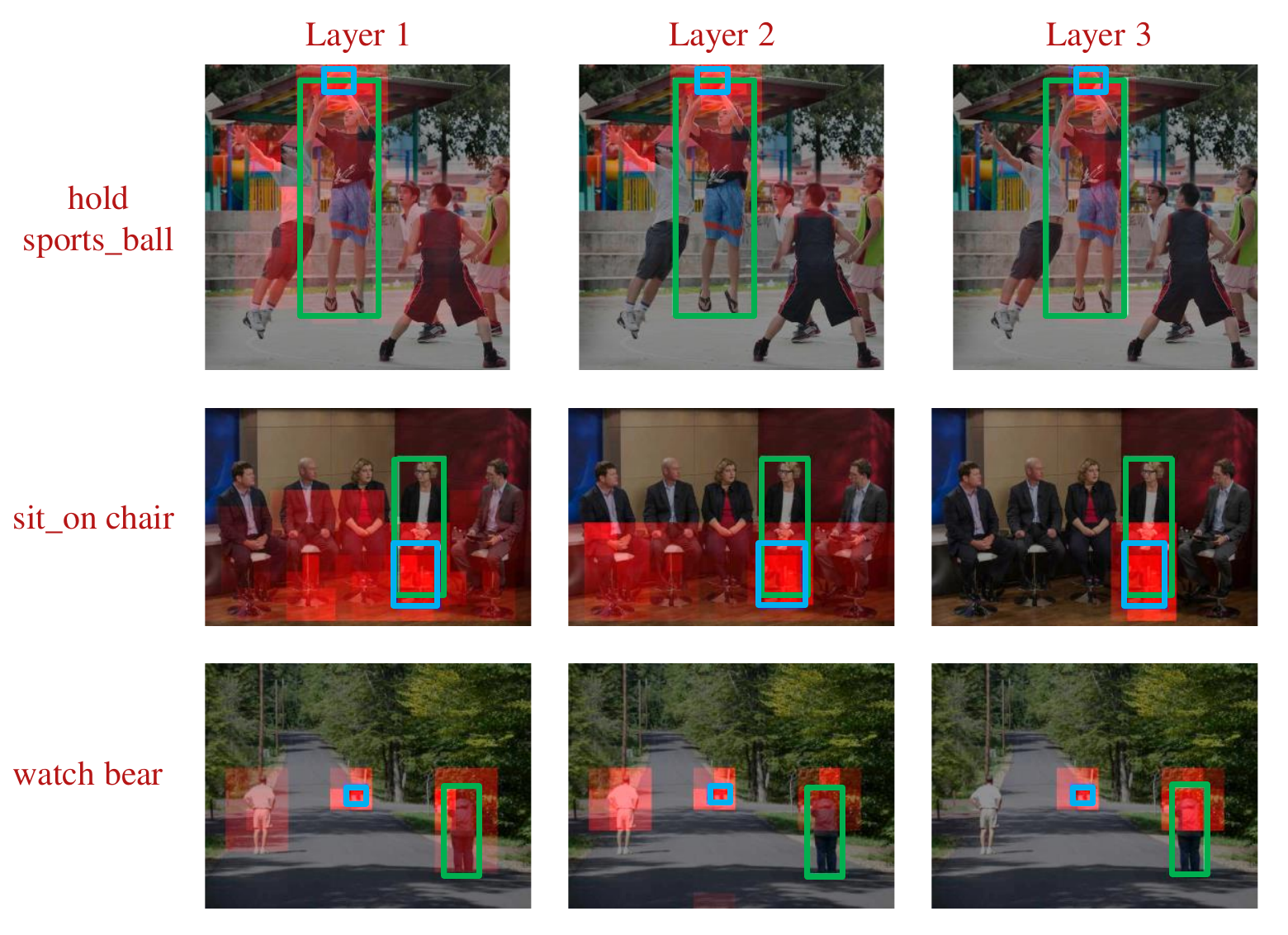}
\caption{More visualization results of how the learned attention changes in each layer, which shows the effectiveness of progressively masking and attention concentration.}
\label{fig:vis_more}
\end{figure}

\section{Visualization}
In Fig.~\ref{fig:examples}, we extend Fig.~1b and show some examples where our proposed global perspective facilitates interactiveness learning. For example, in the left-most image of the upper row, the athlete is inspecting a sports\_ball, and her competitor who is inspecting and reaching for it can provide some useful information. 

Fig.~4 shows some visualization results of the learned attention. Here we provide a detailed analysis.
In Fig.~4d, f, our model learns the attention on the same body-parts (head \& hands) of the target person and other persons.
In Fig.~4a/c, when classifying hands/hands interactiveness of the target person, our model learns to highlight arms \& legs/legs \& feet of other persons to explore visual cues.
For hard cases, train passengers (Fig.~4e) and other athletes in the field (Fig.~4g) provide useful visual cues.

Moreover, Fig.~\ref{fig:vis_more} shows more visualization results of how the learned attention changes in each layer. The 1-st layer allows attention computation from different body-parts of different persons. The 2-nd layer emphasizes the same body-part from different persons in the image, while the 3-rd layer focuses on the body-part of the targeted person. We can see that with the progressive masks throughout transformer layers, different visual patterns are flexibly encoded to facilitate body-part interactiveness learning.

\begin{figure}
\centering
\includegraphics[width=0.95\textwidth]{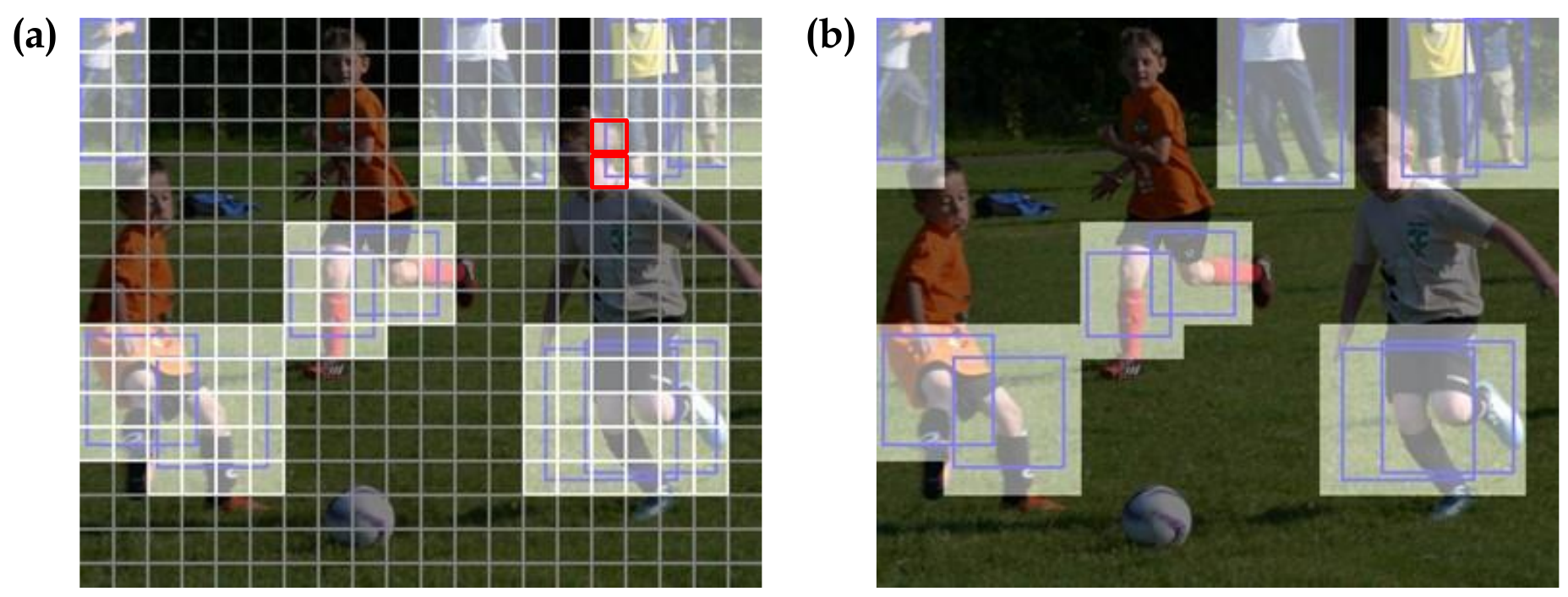}
\caption{The resolution of the feature map, the detected body-part boxes and the calculated attention masks. a) The resolution of the feature map (the grid). b) The detected body-part boxes (left/right legs in blue boxes) and the calculated attention masks (highlighted white regions).}
\label{fig:masks}
\end{figure}

\section{Detailed Design of Attention Masks}
In Sec.~3.2 of the main text, we propose to construct body-part saliency maps via transformer attention masks.
In this section, we introduce the detailed design.

Given a transformer layer in $f_{dec2}$, the input queries are $D=\{d^i| d^i \in \mathcal{R}^{D_c}\}_{i=1}^{N_q}$, and the keys and values $K, V \in \mathcal{R}^{S \times D_c}$ ($S=H \cdot W$) are obtained from feature map $z$. For the $i$-th proposal, originally we have the self-attention calculation as $Att(d^i,K,V) = softmax(d_i K^{T}/\sqrt{D_c})V$. With a mask matrix $m^i \in \{0,1\}^{S}$(0 for masked) to emphasize body-parts, 
we have $Att^{*}(d^i,K,V) = softmax(m^{i*} \circ (d^i  K^{T})/\sqrt{D_c})V$, where 
$\circ$ is Hadamard product,
$m^{i*}=\{m^{i*}_s|m_{s}^{i*}=m^{i}_s(m^{i}_s=1) \ or \ m_{s}^{i*}=-inf(m_{s}^i=0)\}$ and $inf$ is numerically big enough (\textit{e.g.}, $2^{32}-1$). Thus, unrelated tokens are dropped from self-attention calculation.

For the attention mask in Eq.~1, the resolution of the feature map $z$ is scaled down from that of the original image and the scaling factor is 32 with ResNet-50 backbone. Masks are applied on the feature map instead of the original image for the convenience of model design. 
Despite the down-sampling, masks on the feature map can accurately express body-parts. An example is shown in Fig.~\ref{fig:masks}. For each of the three boys, their legs occupy more than 15 tokens. Even for the background person in the left upper corner, he occupies 10 tokens.

Another consideration is that, for the tokens near the body-part boxes border, they may have a small overlap with the useful body-part while a large overlap with the background, thus bringing noisy information, or a large overlap with other body-parts, thus bringing confusion. 
Thus, we propose to randomly drop these tokens based on how much ratio a token is inside the part box. For example, the two red tokens in Fig.~\ref{fig:masks}a may be randomly dropped. Nevertheless, we find the detection performance is slightly changed (interactiveness AP is 38.72 /38.74 w/o random dropping). Therefore, the mask is accurate enough to express the semantics of body-parts despite some geometrical ambiguity because most of the tokens contain the correct body-parts.

\begin{figure}
\centering
\includegraphics[width=0.7\textwidth]{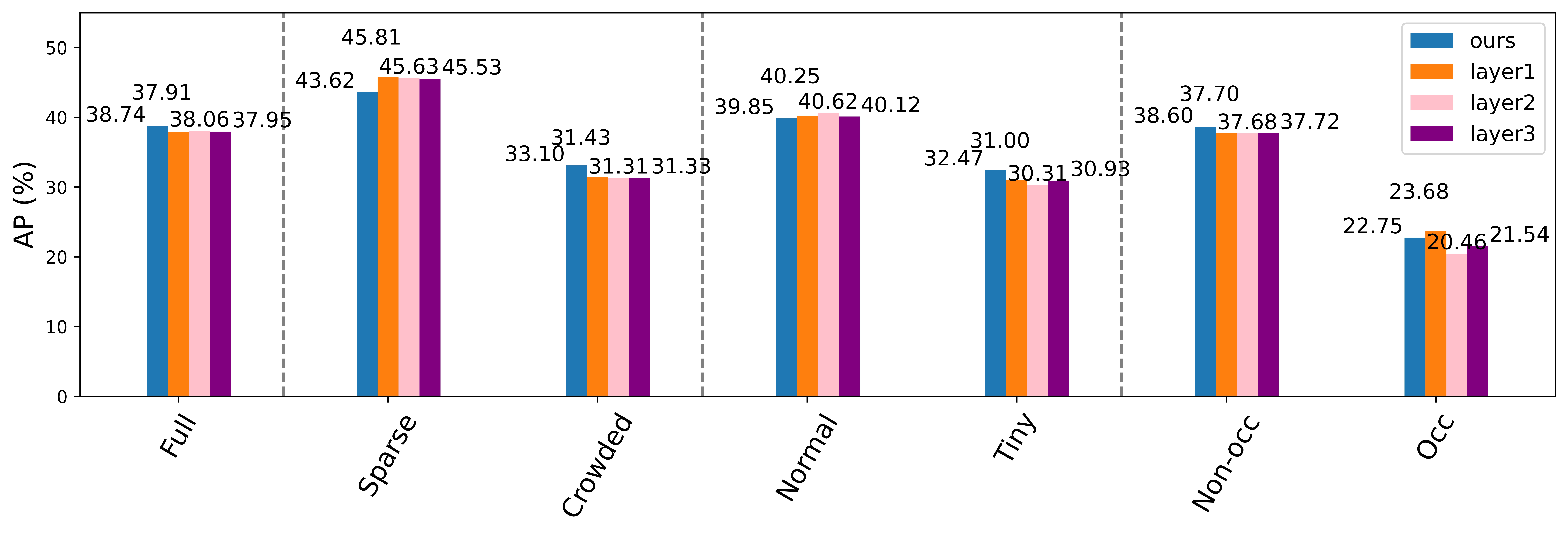}
\caption{Interactiveness detection AP on hard cases with progressively masking removed.}
\label{fig:suppl_ap}
\end{figure}

\section{Analysis of Progressively Masking}
In this section, we analyze the effect of progressively masking in detail. 
As an extension of the ablation studies in Sec.~5.5 of the main text, we remove progressively masking and apply the same attention mask on all decoder layers in $f_{dec2}$ when classifying interactiveness. The interactiveness AP is 37.91/38.06/37.95 when applying masks $m_1/m_2/m_3$ on all layers, while AP is 38.74 with progressively masking. Additionally, we report the detailed performance on hard cases in Fig.~\ref{fig:suppl_ap}. Progressively masking has a slight advantage for sparse scenes with only one/two interactive pairs, while an obvious advantage for other cases, especially for hard cases. The results validate its effectiveness to integrate diverse body-part oriented visual patterns flexibly.

\section{Discussion of Limitations}
Despite the effectiveness of the proposed global perspective, there are still some limitations.
One limitation is the accuracy of body-part boxes from pose estimation results. It may be better to integrate
body-part box regression into the training. Moreover, body-part interactiveness would be learned better if visual patterns across different images are considered.
We plan to improve them in the future work.

\clearpage
\bibliographystyle{splncs04}
\bibliography{egbib}
\end{document}